\documentclass[letterpaper]{article} 
\usepackage{aaai2026}  
\usepackage{times}  
\usepackage{helvet}  
\usepackage{courier}  
\usepackage[hyphens]{url}  
\usepackage{graphicx} 
\usepackage{natbib}  
\usepackage{caption} 
\usepackage{amsfonts} 
\usepackage{algorithm}
\usepackage{algorithmic}

\usepackage{newfloat}
\usepackage{listings}

\usepackage{amsmath}
\usepackage{array} 

\usepackage{tikz}
\usetikzlibrary{arrows.meta,positioning,calc}

\newtheorem{definition}{Definition}

\DeclareCaptionStyle{ruled}{labelfont=normalfont,labelsep=colon,strut=off} 
\floatstyle{ruled}
\newfloat{listing}{tb}{lst}{}
\floatname{listing}{Listing}
\title{Epistemology Gives a Future to Complementarity in Human-AI Interactions}

\author{
    Andrea Ferrario\textsuperscript{\rm 1,2,3}\thanks{Corresponding author: \texttt{aferrario@ethz.ch}},
    Alessandro Facchini\textsuperscript{\rm 2,4},
    Juan M. Dur\'an\textsuperscript{\rm 5}
}

\affiliations{
    \textsuperscript{\rm 1}Institute of Biomedical Ethics and History of Medicine, University of Z\"urich, Z\"urich, Switzerland\\
    \textsuperscript{\rm 2}SUPSI, Dalle Molle Institute for Artificial Intelligence (IDSIA), Lugano, Switzerland\\
    \textsuperscript{\rm 3}ETH Z\"urich, Z\"urich, Switzerland\\
    \textsuperscript{\rm 4}Management in Networked and Digital Societies (MINDS) Department, Kozminski University, Warsaw, Poland\\
    \textsuperscript{\rm 5}TU Delft, Delft, The Netherlands
}

\usepackage{bibentry}

\begin{document}

\maketitle

\begin{abstract}
\emph{Human-AI complementarity} is the claim that a human supported by an AI system can outperform either alone in a decision-making process. Since its introduction in the human-AI interaction literature, it has gained traction by generalizing the reliance paradigm and by offering a more practical alternative to the contested construct of trust in AI. Yet complementarity faces key theoretical challenges: it lacks precise theoretical anchoring, it is formalized only as a post hoc indicator of relative predictive accuracy, it remains silent about other desiderata of human-AI interactions, and it abstracts away from the magnitude-cost profile of its performance gain. As a result, complementarity is difficult to obtain  in empirical settings.
In this work, we leverage epistemology to address these challenges by reframing complementarity within the discourse on \emph{justificatory AI}. Drawing on computational reliabilism, we argue that historical instances of complementarity function as evidence that a given human-AI interaction is a reliable epistemic process for a given predictive task. 
Together with other reliability indicators assessing the alignment of the human-AI team with the epistemic standards and socio-technical practices, complementarity contributes to the degree of reliability of human-AI teams when generating predictions. 
This theoretical repositioning supports the practical reasoning of those affected by these outputs---patients, managers, regulators, and others. Our approach suggests that the role and value of complementarity lie not in providing a stand-alone measure of relative predictive accuracy, but in helping calibrate decision-making to the reliability of AI-supported processes. We conclude by translating this epistemological repositioning into design- and governance-oriented recommendations, including a minimal reporting checklist for justificatory human-AI interactions and measures of efficient complementarity.
\end{abstract}

\section{Introduction}
AI-assisted decision-making is increasingly used in high-stakes domains such as healthcare, education, and public administration. In such settings, a central question is whether interaction between a human and an AI system actually improves decision quality relative to either acting alone. Research on human-AI interaction captures this expectation under the label \emph{complementarity}: in prediction tasks, a human assisted by an AI system, often referred to as a \emph{human-AI team}, should outperform both the human and the AI alone \citep{Bansal2021CHI,Bansal2021AAAI}.
Since \citet{Bansal2021CHI}, complementarity has attracted considerable attention in the human-AI interaction community \citep{Bansal2021AAAI,Bansal2021CHI,hemmer2021human,Hemmer2025EJIS,Vaccaro2024NHB,gonzalez2026toward}. A growing body of work shows that, while humans and AI models often make different kinds of errors \citep{geirhos2021partial},  
human-AI teams can outperform humans alone \citep{alufaisan2021does,inkpen2023advancing,Vaccaro2024NHB}, and, in some settings, they can even outperform AI systems \citep{dvijotham2023enhancing,ma2023should}. Here, complementarity serves  as a design standard that motivates interaction protocols extending the  reliance paradigm, in which the output of the human-AI interaction is constrained to equal either the human’s prediction (self-reliance) or the AI’s prediction (AI reliance) \citep{schemmer_appr_reliance,zhang2020effect,Bansal2021CHI}, by admitting interaction outputs that differ from the human and the AI's predictions as the result, for instance, of repeated exchange of information. 

Despite its intuitive appeal, complementarity faces four key theoretical  challenges. First, complementarity is defined in terms of a performance measure called \emph{complementarity team performance} (CTP)  and its
theoretical anchoring remains limited \citep{hemmer2021human,Hemmer2025EJIS,Donahue2022FAccT}.  
Second, like other performance metrics in supervised learning, CTP is an ex post, ground-truth-dependent measure: it is not available at decision time and its stability under AI system updates and human learning is poorly understood \citep{Hemmer2025EJIS}. 
Third, complementarity  is defined solely over predictive and \emph{relative} accuracy; 
in many contexts, however, accuracy is only one among several decision-relevant desiderata, such as fairness, robustness, explainability, and resource constraints among others \citep{Vaccaro2024NHB}. Finally, the literature on complementarity currently abstracts from discussing the magnitude of the performance gain achieved by the human-AI team, and the cost of interaction required to obtain it. 
Taken together, these observations suggest that, despite the growing body of work in human-AI interaction, complementarity is not suited to function as a stand-alone, \emph{relative} accuracy-focused gold standard for human-AI collaboration.
Not only do we lack a clear theoretical stance on complementarity, but empirically we still struggle to achieve it \citep{Vaccaro2024NHB}. So, \textbf{what future lies ahead of complementarity in human-AI interactions?}

In this paper, we argue that by reconsidering complementarity through an epistemological lens, we can still assign it a meaningful role in human-AI interaction research despite these limitations. Our central claim is that complementarity should not be treated as a stand-alone gold standard for human-AI collaboration, but as historically grounded evidence that a prediction-task human-AI interaction may constitute a reliable epistemic process for a given task. More specifically, we show that complementarity can contribute to the epistemic justification of outputs produced by human-AI teams.
To do so, we reinterpret it as one reliability indicator within \emph{computational reliabilism}  \citep{duran2018grounds,duran2025defense,duran2026}, a process-reliabilist framework that connects the historical performance and surrounding practices of computational processes to the justification of beliefs formed on their basis. On this view, complementarity contributes to what we call \emph{justificatory AI} \citep{Ferrario2024SEE,alvarado2023ai,alvarado2023kind} by helping answer a practical question for affected third parties: \emph{when is it reasonable for a patient, student, manager, or regulator to accept the output of a human-AI team as epistemically adequate for a given task?}
 Our contributions are as follows: 
\begin{itemize}
\item \textbf{We synthesize the main theoretical challenges to complementarity} identified in the human-AI interaction literature, and discuss how they constrain its empirical effectiveness.

\item \textbf{We reconceptualize complementarity within computational reliabilism}, treating prediction-oriented human-AI interactions as computational processes whose reliability comes in degrees and is supported by heterogeneous markers, with historical instances of complementarity serving as one important indicator. This shows how complementarity can contribute to the epistemic justification of outputs produced by human-AI teams. Furthermore, CR allows clarifying what should be measured and reported beyond relative accuracy, e.g., stability under updates, task validity, and governance and competence scaffolding, to assess epistemic adequacy of human-AI teams.

\item \textbf{We argue that researchers should target \emph{efficient} complementarity}: if predictive gains require sustained monitoring, prolonged deliberation, or extensive retraining, the design may merely shift epistemic burden onto users and erode the practical value of the improvement. To operationalize this perspective, we provide a minimal reporting checklist for \emph{justificatory human-AI interactions} and measures of efficient complementarity.
\end{itemize}


\section{The Basics: Human-AI Interactions, Reliance, and Complementarity}
\label{section:complementarity}
We begin with a minimal formal setup for human-AI interactions, reliance, and complementarity.

\subsection{Human-AI Interactions}
\label{subsection:HAI_int}
We call \emph{human-AI interactions} (HAIs) goal-oriented processes in which a human and an AI system contribute informational inputs that are  combined into a single output, e.g., a prediction or AI-generated content, possibly after update.\footnote{This definition is deliberately minimal: it characterizes HAIs at the level of information exchange and output formation, without committing to any particular interaction protocol, degree of agency, or normative relationship between the agents.} In prediction settings, such as a clinician consulting a medical AI to classify skin lesions, the informational inputs are typically the agents' per-instance predictions: the AI is used as a recommendation-driven decision-support system \citep{miller2023explainable}.\footnote{Miller uses the term \emph{recommendation} for what we call a \emph{prediction} \citep{miller2023explainable}. More generally, however, these AI inputs may include any informational content supplied to the interaction protocol; for instance the AI system may provide  information to help the human refine their prediction \emph{without} sharing a prediction, i.e.,  as a hypothesis-driven decision support \citep{miller2023explainable}.
} 
In this work, interactions whose goal is 
to approximate ground truth for labeled instances play a key role. 

\subsection{Prediction-task-oriented HAIs and the Self-Reliance vs. AI Reliance Paradigm}
\label{subsection:PT-HAI_Reliance}
A prediction task $\tau$ is the datum of an input space $\mathcal X$, an output/label space $\mathcal Y$, and a pointwise loss
$\ell:\mathcal Y\times \mathcal Y \to \mathbb{R}_{\ge 0}$ used to evaluate predictions against ground truth. Given $\tau$, a labeled dataset is a finite sample $D:=\{(x_i,y_i)\}_{i=1}^{n}$ with $x_i\in\mathcal X$ and $y_i\in\mathcal Y$. Typically, datasets are used to estimate performance for $\tau$ of human-AI interactions over time. More precisely,
a \emph{prediction-task human-AI interaction} (PT-HAI)---in symbols: $\text{PT-HAI}^{\tau}(\Pi)$---for  $\tau$ is an HAI whose output $\hat y_i^{HAI}$ is the approximation (or prediction) of the ground truth  $y_i$ of any instance $(x_i,y_i)$ in a dataset $D$.
A $\text{PT-HAI}^{\tau}(\Pi)$ is typically instantiated  over time for the same task $\tau$, under a fixed interaction protocol $\Pi$, i.e., the specification of how the human consults and integrates the AI, but on (possibly different) datasets $D_1,\dots,D_k$, where each $D_j$ is a finite sample from $\mathcal X\times\mathcal Y$. When unambiguous, we omit the explicit reference to $\tau$ and $\Pi$.\footnote{We treat PT-HAIs as processes that can be instantiated many times over time, on different cases and datasets, and potentially by different qualified humans, while keeping the task, protocol, and the AI  fixed. We separate the process from its tokens, i.e., the interaction instances. This matters later, because our reliability claims introduced in Section \ref{section:compl_reliabilism} concern the PT-HAI, not any single token.}
 

If the AI is used as a recommendation-driven decision support, then the inputs of the PT-HAI are the human's and the AI's  predictions \(\hat y_i^{H}\) and \(\hat y_i^{AI}\) of $y_i\in D$ for a given task $\tau$. This is the case for a prominent class of PT-HAIs called \emph{self-reliance vs.\ AI reliance}---in short: \emph{reliance}---where 
no interaction between the predictions of the human and AI occurs, and the output given by the selector of the two single-agent predictions: if \(\hat y_i^{HAI}=\hat y_i^{H}\) we have \emph{self-reliance}; if \(\hat y_i^{HAI}=\hat y_i^{AI}\) we have \emph{AI reliance} \citep{schemmer_appr_reliance,zhang2020effect,ferrario2025being} instead. We use ``reliance'' in the narrow selector sense, to keep it contrastive with interactions where information exchange between the human and the AI allows outputs outside $\{\hat y^H,\hat y^{AI}\}$---see Section \ref{subsection:complementarity} below.
If the human is initially wrong, the AI is right, and the human follows the AI or the human is initially right and ignores incorrect AI advice, then \emph{appropriate reliance} emerges. This is  the optimal target of reliance relations and it is actively discussed in the human-AI interaction and philosophy of AI literature \citep{schemmer_appr_reliance,zhang2020effect,ferrario2025being}.

\subsection{Complementarity}
\label{subsection:complementarity}
\citet{Bansal2021AAAI} observe that, in many contexts where humans use AI systems in decision-making, we are interested in the performance of the human-AI team rather than that of either component alone. The reason is that, in many applications, AI systems provide \emph{advice} that humans incorporate into decision-making.
(They therefore exclude cases of full automation.) In these scenarios, the AI system assists humans by sharing its outputs, while human agents retain accountability on the final decisions \citep{Bansal2021CHI}: these decisions are taken by the human assisted by the AI or, hereafter, the human-AI team.\footnote{
In this work, human-AI team refers to the human agent responsible for producing a prediction as a result of their interaction with an AI system. Our terminology is aligned with what \citet{Hemmer2025EJIS} define as ``human-AI collaboration.''}
Paradigmatic cases happen in high-stakes domains, such as medicine and the judicial system where experts, e.g., doctors and forensic experts, use AI systems to assist their decision-making, e.g., diagnoses, as well as facial and voice detection assessments. Following \citep{Hemmer2025EJIS,Donahue2022FAccT}, we introduce complementarity as follows: 

\begin{definition}[Complementarity]\label{def:complementarity}
Let $\text{PT-HAI}^{\tau}(\Pi)$ be a prediction-task human-AI interaction with protocol $\Pi$, relative to a task $\tau$ and dataset $D:=\{(x_i,y_i)\}_{i=1}^n$, with human and AI predictions $\hat{y}^H_i$ and $\hat{y}^{AI}_i$ for each $y_i$, and a team output $\hat{y}^{HAI}_i$.
Let $L_H$, $L_{AI}$, and $L_{HAI}$ be empirical losses\footnote{These  aggregate the loss $\ell$ computed at each prediction. They depend on the prediction task. 
} measuring the error incurred by the human, the AI system, and the human-AI team, respectively, in approximating the ground truths $y_i$ in $D$. 

Define the \emph{Complementarity Team Performance} for task $\tau$ relative to $D$ by:
\begin{equation}
\text{CTP}_\tau(D):=
\begin{cases}
1, & \text{if } L_{HAI}(D) < \min\{L_{H}(D),L_{AI}(D)\},\\[4pt]
0, & \text{otherwise}.
\end{cases}
\label{eq:CTP}
\end{equation}

We say that $\text{PT-HAI}^{\tau}(\Pi)$ \emph{achieves complementarity on $D$} whenever $
\text{CTP}_\tau(D)=1$.
\end{definition}

\textbf{Complementarity is a binary property of a PT-HAI}: a human-AI interaction achieves complementarity,\footnote{Note that \citet{Vaccaro2024NHB} calls complementarity ``human-AI synergy,'' that is ``where the human-AI group performs better than both the human alone and the AI alone'' \citep[pag.~2294]{Vaccaro2024NHB}. We refer to \citep{Hemmer2025EJIS},  for a review of concepts similar to complementarity that commonly appear in the literature.}  i.e., $\text{CTP}_\tau(D)=1$, as per eq. \eqref{eq:CTP}, when the human and the AI use their capabilities in such a way that their respective errors are mitigated or corrected \citep{Bansal2021AAAI}.
Importantly, as remarked by \citet{Hemmer2025EJIS}, the human-AI interaction output $\hat{y}^{HAI}_i$ may be different from $\hat{y}^H_i$ and $\hat{y}^{AI}_i$. 
Figure \ref{fig:HAI_int_comple} depicts human-AI interactions fitting the complementarity setting. Two remarks. First, Def~\ref{def:complementarity} can be generalized in terms of objectives, such that utilities, payoffs or  policies, that may differ from empirical losses \citep{rastogi2023taxonomy,Donahue2022FAccT}, but this does not affect our core argument. Similarly, the binary measure $\text{CTP}_\tau(\cdot)$ can be relaxed and replaced by continuous ones; also this choice does not affect our core argument. Furthermore, complementarity extends reliance by requiring active interaction that entails the exchange of information between the human and the AI system, resulting in a human-AI team's prediction that does not need to match either the human's or AI's. For instance, in regression tasks, the human-AI team may average or otherwise combine the two predictions as the result of this informational exchange. In multiclass classification, the human-AI team may select a class that neither the human nor the AI initially (and independently) proposed. As a result, this collaboration augments the space of outputs of the interaction.\footnote{PT-HAIs where the AI is used purely as an information source to help the human refine their initial prediction can still be analyzed through the lens of complementarity, provided the AI is in principle capable of computing predictions.}

\begin{figure*}[t]
  \centering

  \begin{tikzpicture}[
  scale=0.85,
    transform shape,
      >=Latex,
      every node/.style={font=\small},
      box/.style={draw, minimum width=0.7cm, minimum height=1.1cm, align=center},
      circ/.style={draw, circle, minimum size=9mm, inner sep=0pt}
    ]

    \node (IX) at (0,0.4) {$\hat{y}^{H}_i$};
    \node (IS) at (0,-0.4) {$\hat{y}^{AI}_i$};

    \node[box] (phi) at (4.0,0) {};

    \node[circ] (Gamma) at (9.0,0) {};

    \draw[->] (IX.east) -- ($(phi.west)+(0,0.4)$);
    \draw[->] (IS.east) -- ($(phi.west)+(0,-0.4)$);

    \path let \p1 = ($(phi.east)+(0,0.4)$) in coordinate (A) at (\x1,\y1);
    \draw[->]
      (A) -- ++(1.6,0)
      to[out=0,in=140] (Gamma.120)
      node[pos=0.45, above] {$\hat{y}^{H'}_i$};   

    \path let \p1 = ($(phi.east)+(0,-0.4)$) in coordinate (B) at (\x1,\y1);
    \draw[->]
      (B) -- ++(1.6,0)
      to[out=0,in=220] (Gamma.240)
      node[pos=0.45, below] {$\hat{y}^{AI'}_i$};  

    \draw[->] (Gamma) -- ++(2.2,0) node[right] {$\hat{y}^{HAI}_i$};

    \node[above=1pt of IX] {Input (human)};
    \node[below=1pt of IS] {Input (AI)};
    \node[below=10pt of Gamma] {Output};
    \node at ($(phi.south)+(0,-0.45)$) {Input interaction};
  \end{tikzpicture}
  \caption{A graphical depiction of human-AI interactions that fit the complementarity setting. The goal of the interaction is to arrive at ``human-AI team'' predictions, given a prediction task $\tau$ and a dataset $D=\{(x_i,y_i)\}_{i=1}^n$. 
  The human and AI inputs are updated throughout the interaction and complementarity is reached if \eqref{eq:CTP} holds. In case of reliance interactions, the  interaction is trivial, namely, 
  $\hat{y}^{H'}_i=\hat{y}^{H}_i$, $\hat{y}^{AI'}_i=\hat{y}^{AI}_i$, and the output results from an input selector:  
  $\hat{y}^{HAI}_i\in \{ \hat{y}^{H}_i, \hat{y}^{AI}_i \}$ for all $i=1,\dots,n$.
  }
  \label{fig:HAI_int_comple}
\end{figure*}


Complementarity remains attractive in human-AI research because it shifts attention from optimizing AI systems in isolation to understanding when collaboration between humans and AI systems yields genuine epistemic gain in practice \citep{miller2023explainable,Hemmer2025EJIS}. Furthermore, $\text{CTP}_\tau(\cdot)$ provides an easy-to-understand quantitative measure of human-AI collaboration, helping shift attention away from contested constructs such as trust in AI and calibrating trust in trustworthiness, which have been widely criticized in both philosophy and human-AI interaction \citep{ryan2020ai,duran2025trust,Benk2025AISoc,ferrario2025being}. That said, as currently formulated, complementarity faces several key theoretical challenges, which we discuss below.

\subsection{Complementarity: Key Theoretical Challenges}
\label{subsection:compl_challenges}
We list four theoretical challenges affecting complementarity that will motivate our approach in Section \ref{section:compl_reliabilism}.



\vspace{1em}
\noindent\textbf{I. What theoretical anchoring for complementarity?} \emph{Complementarity still lacks a clear theoretical positioning}. \citet{Hemmer2025EJIS} suggest that empirical findings showing humans and AI systems making different types of errors can be framed within an Input-Processing-Output model of decision-making \citep{d2014reflecting},  which rests on two key ideas: information and capability asymmetry. However, the model is highly general, used across domains and technological artifacts, and does not distinguish between different kinds of human-AI interactions, such as reliance versus complementarity. It highlights differences in information and its processing, but not the concrete mechanisms by which inputs are compared, reconciled, and integrated into a human-AI team decision, see Fig. 1 in \citep{Hemmer2025EJIS}. 
A more recent and notable attempt is the framework by \citet{gonzalez2026toward}, which offers a sociotechnical vocabulary for thinking about human-AI teaming across dimensions such as reasoning, memory, attention, and governance. Yet this framework also remains high-level and programmatic: it is neither formal nor prescriptive, and its empirical basis is still partly limited by the early-stage and heterogeneous nature of current evidence.

\vspace{1em}
\noindent\textbf{II. Limited applicability of complementarity at decision time.} 
CTP and other complementarity-related metrics, see \citet{Hemmer2025EJIS}, depend on ground truth labels and summarize historical performance over a set of cases, epistemically analogous to accuracy or mean square error in supervised learning.\footnote{Practically, however, complementarity differs from standard supervised-learning evaluation as in many applied pipelines collecting labels to estimate accuracy-type metrics is easy to moderately difficult: labels can be obtained through routine outcomes, or periodic audits. By contrast, estimating complementarity additionally requires  instrumenting and recording the human-AI interaction itself: one must capture not only ground truth $y_i$, but also the human's and AI's predictions (possibly multiple, updated over the interaction), the interaction protocol, and the team output $\hat y_i^{HAI}$ in a way that supports faithful reconstruction of the process.} They therefore do not  tell a decision-maker at a given time, in the absence of ground truth, whether to rely on the AI, on themselves, or on an alternative information source (and if ground truth were
available at decision time, AI predictions would be redundant). In other words, we have no direct way to measure the emergence (or absence) of complementarity as decisions are being made:  \emph{complementarity is  an ex post construct to evaluate human-AI interactions}.\footnote{Note that Alur, Raghavan, and Shah's framework \citeyearpar{alur2024human} can guide design ex ante, but their
``superiority relations'' are still learned against post-hoc labeled data (see their Algorithm 1).} Finally, current formulations of complementarity are silent on how robust
CTP must be across temporal changes, e.g., data distribution shifts and human epistemic process update, and under what conditions one may expect complementarity to persist \citep{Hemmer2025EJIS,Vaccaro2024NHB}.

\vspace{1em}
\noindent\textbf{III. More than \emph{relative} predictive accuracy is at stake in human-AI interactions.} 
Complementarity is a \emph{relative} accuracy criterion: a human-AI team can satisfy CTP even if its absolute performance is very low, as long as it is better than each component alone: even very unskilled humans and highly inaccurate AIs can still achieve complementarity—simply by being less bad together than each is on their own. However, many real-world decision settings impose constraints that are not captured by relative loss comparisons alone: robustness to distribution shift, validity of the target construct, transparency and fairness requirements, resource constraints, legal compliance, and institutional accountability, among others. Some of these desiderata are not purely epistemic, but in high-stakes contexts they function as practical constraints on what counts as an epistemically acceptable decision procedure. 

A human-AI team can display excellent aggregate predictive performance while simultaneously amplifying disparate error rates across subgroups  \citep{Donahue2022FAccT}. Because CTP is computed at the dataset level, subgroup-level complementarity failures can be masked by aggregate gains, a structural limitation that any stand-alone use of CTP inherits. More broadly, a team can satisfy CTP while increasing cognitive burden on the human agent through excessive monitoring of the AI system and its outputs, or while undermining accountability and patient trust. 

Thus, \emph{complementarity is, at best, a conditional desideratum of human-AI interactions}: it is epistemically and practically attractive only when it is achieved 
subject to constraints on other epistemic or non-epistemic (e.g., ethical) values.

\vspace{1em}
\noindent\textbf{IV. Complementarity ignores the magnitude-cost profile of epistemic gain.}
Beyond being a relative accuracy criterion, complementarity abstracts away from two relevant  dimensions: (1) the magnitude of the performance gain achieved by the human-AI team, and (2) the cost of interaction required to obtain it. In practice, the epistemic and practical significance of complementarity gains varies widely. In some settings, achieving CTP yields only negligible improvements over reliance; in others, it results in substantial error reduction.
However, achieving complementarity may require non-trivial costs, including increased cognitive effort, longer decision times, additional training and monitoring, or institutional commitments. The current literature does not account for this gain-cost trade-off: as a result, \emph{complementarity treats interactions with very different epistemic and practical profiles as equivalent, despite their divergent implications for real-world deployment}.

\vspace{1em}

These theoretical challenges affect complementarity in empirical settings. In fact, research indicates that, although there is some partial support, complementarity does not emerge in practice systematically. Vaccaro et al.'s \citep{Vaccaro2024NHB} meta-analysis of 100+ studies shows that, while human-AI teams frequently outperform humans alone,  on average, human-AI teams do not perform better than either humans or AI alone. 
It also shows that task type  moderates complementarity: for prediction tasks the pooled effect size for complementarity is significantly negative, whereas it is positive for creation tasks, such as AI-assisted generation of artistic images or text. Design interventions such as explanations or confidence displays do not reliably induce complementarity and  hindrances include miscalibrated self-assessment, coordination frictions, and forms of error mirroring 
\citep{turel2023prejudiced,Vaccaro2024NHB,vaccaro2019effects,zhang2020effect,cabitza2021studying}.

\vspace{1em}

Given these theoretical limitations and empirical challenges, complementarity should be theoretically repositioned in AI-assisted decision-making. Our claim is not that complementarity lacks value. Rather, when treated as an ex post, ground-truth-dependent, relative-accuracy criterion that is difficult to obtain empirically, it is poorly suited to function as a stand-alone gold standard for human-AI interaction. What is needed is a theoretical repositioning that preserves what is informative about complementarity while embedding it in a broader account of epistemic adequacy. We defend this perspective by positioning complementarity within a framework for justifying the predictions of humans assisted by AI systems. Our approach is described next.


\section{Our Approach: Complementarity and Computational Reliabilism}
\label{section:compl_reliabilism}
Here, our aim is to show that complementarity can play a meaningful role in the epistemic justification of outputs produced by human-AI teams, reinterpreting it as one reliability indicator within the framework of computational reliabilism (CR) \citep{duran2018grounds,duran2024understanding,duran2025defense}. 
To do so, we need to start with a primer on CR.

\subsection{A Primer on Computational Reliabilism}
\label{subsection:CR_primer}
Goldman's \emph{process reliabilism} holds that a belief is justified when it is produced by a reliable belief-forming process, that is, one that tends to yield true (or otherwise epistemically adequate) outputs in the relevant circumstances \citep{Goldman1986Book}. Computational reliabilism applies and extends this idea to cases in which beliefs are formed through direct or indirect interaction with \emph{computational processes}, including those realized through computer simulations, algorithms, and AI systems \citep{duran2025defense,duran2018grounds,duran2026}. Thus, on CR, the central unit of epistemic evaluation is the computational process that performs a task $\tau$, e.g., generating a prediction, and the justificatory question is whether that process is \emph{reliable} in a given context---we explain the notion of reliability below. Here, a computational process is an input-output procedure that transforms information according to some method, e.g., statistical or rule-based inference, or  simulation. The computational process is embedded in a socio-techno-scientific context that shapes the design of the entity, e.g., an AI system, realizing the computation at different instantiations (or tokens) of the process over time  \citep{duran2025defense, duran2026}. 
CR reads:

\begin{definition}[Computational reliabilism]
\label{def:CR_algo}
Let \(X\) be an agent and \(A\) a computational process used for task \(\tau\). \(X\) is justified in accepting \(A\)'s output as epistemically adequate for \(\tau\) iff \(A\) is \emph{reliable} for \(\tau\).
\end{definition}

In Def. \ref{def:CR_algo}, reliability  is dispositional: a computational process is \emph{reliable} when, across the kinds of cases in which it is used for $\tau$, it \emph{usually} yields epistemically adequate results (for instance, medically valid predictions in a medical context by an AI system) \citep{duran2025defense, duran2026}. Here, \emph{epistemically adequate} is a domain-relative success notion, see \citep{duran2025defense, duran2026}. In prediction tasks, adequacy is primarily predictive accuracy  under the domain-relevant performance measure, but in high-stakes it includes also 
disciplined treatment of uncertainty and adherence to constitutive domain constraints, e.g., scope of use and acceptable error trade-offs. 
The reliability of the computational process provides $X$ with
defeasible reasons to treat outputs as admissible inputs to action and further deliberation.
CR is compatible with both frequentist and propensity-style interpretations of \emph{usually}; we will not (need to) take a stand on this debate here.\footnote{For classic discussions, see  \citet{alston1995think}.} 

Crucially, under CR, the reliability of the computational process is given by families of properties, capabilities, and practices that provide defeasible evidence that the process is reliable in the context where it is performed. These properties, capabilities, and practices are collected into  \emph{reliability indicators}, understood as marks of methodological, cognitive, social, and epistemic competence. Thus, endorsing CR, the identification of reliability indicators is key to justifying $X$'s belief that the output of a computational process is epistemically adequate. More precisely, CR distinguishes between categories of indicators (\texttt{type-RI}s), whose  content depends on what entity realizes the computational process:

\begin{enumerate}
\item \textbf{the technical performance and operational behavior of the realizing entity} (\texttt{type-RI$_1$});
\item \textbf{the proper operationalization of scientific concepts, models, and domain knowledge} (\texttt{type-RI$_2$});
\item \textbf{the social construction of reliability} (\texttt{type-RI$_3$}).
\end{enumerate}

As a brief example, consider a computational process realized by an AI system (e.g., fracture detection from medical images or sepsis prediction from physiological streams). \texttt{type-RI$_1$} captures the system’s engineering and operational performance, namely, how it is built and maintained, and how well it behaves in use, e.g., accuracy, error modes, robustness to shifts  \citep{duran2025defense, duran2026}. \texttt{type-RI$_2$} captures epistemic and scientific fit: how domain concepts, models, and values from bioinformatics and medicine are translated into computational form, and how well that operationalization supports inquiry in everyday clinical practice \citep{duran2024understanding,duran2025defense, duran2026}. 
Finally, \texttt{type-RI$_3$} is concerned with how different communities, e.g., bioinformaticians, clinicians, nurses, and the public come to endorse or dismiss AI systems and their outputs, how well these systems realize their intended values and purposes, and how their overall epistemic standing is judged. \texttt{type-RI$_3$} indicators concern institutionalized practices that make reliability contestable, auditable, and corrigible over time, e.g., public and expert debate, stress-testing, and other forms of critical, collective scrutiny \citep{duran2024understanding,duran2025defense, duran2026}.

Two remarks are due. First, while CR provides justification for predictions of computational processes, these predictions can  occasionally be imprecise, inadequate for a specific purpose, or simply wrong. Over time, the accrual of these errors would affect the propensity of the process to compute usually epistemically adequate predictions, ultimately making it unreliable. Furthermore, although conceptually distinct, the reliability indicators interlock in practice and jointly support the process reliability. In fact, the reliability of a computational process comes in degrees and the weight a particular type of reliability indicator carries for a process's reliability depends on the socio-technical context in which it unfolds. The relative importance of a reliability indicator type  is shaped by the epistemic and practical aims in play, as well as by the values of the relevant epistemic community, namely, their habits and norms shaping the management of computational processes. In fact, no single reliability indicator can secure reliability across all processes and contexts (and to the same degree). Even when a given indicator is well-suited to a particular process at one point in time, subsequent developments may undermine its weight: indicators that once seemed compelling can lose their justificatory power as new methods, standards, or evidence emerge \citep{duran2024understanding,duran2025defense, duran2026}.

\subsection{The Role of Complementarity in CR}
\label{subsection:compl_CR}
Finally, we show how complementarity fits within CR. The key move is to treat PT-HAIs as computational processes, and any human-AI team in a given use episode as a token (instantiation) of that process---Section~\ref{subsection:PT-HAI_Reliance}. On this view, the relevant computation is not performed by the AI system alone as in Section \ref{subsection:CR_primer}: it is jointly realized by a human and an AI system acting under a fixed interaction protocol. Two philosophical considerations support this process view. First, Extended Mind Theory and distributed cognition approaches hold that cognitive work can be functionally distributed across people and cognitive artifacts \citep{hutchins1995cognition,hollan2000distributed,clark1998extended,heersmink2015dimensions}. Second, following \citet{alvarado2023ai}, AI systems are epistemic technologies: they can enhance human inquiry by contributing information and capabilities that extend a human user's cognitive and epistemic capabilities. Taken together, these perspectives motivate treating the PT-HAI as a legitimate unit of epistemic assessment in CR and human-AI teams as its token realizations over time. Then:

\begin{definition}[Computational reliabilism for PT-HAIs]
\label{def:CR_hybrid}
Let \(X\) be a human agent, \(A\) an AI system, and let \(\hat{y}^{HAI}\) denote the output of a prediction-task interaction between them for task \(\tau\) and relative to a dataset $D$. A third party \(Z\) (e.g., a patient, student, manager, or regulator) is justified in believing that \(\hat{y}^{HAI}\) is epistemically adequate for \(\tau\) iff \(\hat{y}^{HAI}\) is produced by a \emph{reliable} prediction-task human-AI interaction for \(\tau\).
\end{definition}

Following Section \ref{subsection:CR_primer}, also PT-HAI reliability in Def. \ref{def:CR_hybrid} comes in degrees and it is granted by  reliability indicators.

\vspace{1em}

\noindent \texttt{type-RI$_1$} \emph{and complementarity.} \texttt{type-RI$_1$} indicators capture properties of the PT-HAI as an epistemic process, rather than of the human or the AI in isolation. 
\textbf{Within} \texttt{type-RI$_1$}, \textbf{we treat complementarity---as captured by}  $\text{CTP}_\tau(\cdot)$---\textbf{as a central reliability indicator of the interaction}: it tells us whether, in a given prediction task, the human-AI team performs better than either component alone. Other indicators in \texttt{type-RI$_1$} include systematic patterns of error mitigation or correction, e.g., humans appropriately contest or override AI outputs, escalation and review behavior in difficult cases, stability of joint performance, and the extent to which interaction protocols prevent over- or under-reliance. It also includes disaggregated performance evidence, namely the distribution of joint accuracy and error rates across operationally relevant subgroups, since aggregate complementarity can mask subgroup-level failures that affect the reliability profile of the PT-HAI.
Epistemically, \emph{historical complementarity} supports a counterfactual claim: that had either component acted alone, the PT-HAI would have been \emph{less reliable}. Complementarity does not merely track performance at a given instantiation of a PT-HAI but evidences a division of epistemic labor whose relevance accrues over time. 

That said, \emph{is complementarity necessary for PT-HAI reliability?} In general, the answer is no due to degenerate cases. For instance, suppose the AI system is perfect on task $\tau$ for multiple datasets $D_1,\dots,D_m$, i.e., $L_{AI}(D_k)=0$, $k=1,\dots,m$. Then no PT-HAI can strictly outperform the AI, so $\text{CTP}_\tau(D_k)=1$ is impossible; yet a trivial interaction in which the human simply defers to the AI can still be---to an extent that depends, in particular, on the context---reliable for $\tau$. Similar considerations hold in case of a human with perfect performance on certain datasets of a task $\tau$.\footnote{
One could avoid the degenerate cases by adopting a tolerance-based notion of complementarity, e.g., setting $\text{CTP}_\tau(D)=1$ whenever $L_{HAI}(D)\le \min\{L_H(D),L_{AI}(D)\}$ (or by requiring an improvement of at least $\varepsilon>0$). In this work, we follow the strict definition introduced by \citet{Hemmer2025EJIS} and related work.} Set these degenerate cases aside, what complementarity does provide is a particularly strong \texttt{type-RI$_1$} signal that the interaction protocol enables genuine \emph{epistemic gain} beyond either component alone.  

Accordingly, we treat complementarity as a central indicator of PT-HAI reliability, but neither as a necessary condition nor as a sufficient one.
Thus, on our account, complementarity is \emph{evidential rather than constitutive}: it strengthens the case that a prediction-task human-AI interaction is reliable, but it does not by itself define or secure that reliability.

\vspace{1em}

\noindent \texttt{type-RI$_2$} \emph{and epistemic standards.} \texttt{type-RI$_2$} indicators capture the extent to which the PT-HAI process is aligned with the epistemic and scientific standards that govern sound inference in the relevant domain. \texttt{type-RI$_2$} asks whether the interaction is the right kind of epistemic procedure for task $\tau$ in context. Concretely, it includes indicators such as: (i) \emph{task and target validity}, i.e., whether $\tau$ is well-defined and whether $\hat{y}^{HAI}$ targets a construct that is epistemically appropriate (rather than a poorly motivated proxy); (ii) \emph{evidential fit}, i.e., whether the inputs and features used by the AI system and the human's background information are relevant and adequate for $\tau$; (iii) \emph{validation and generalization standards}, i.e., whether there is adequate evidence that the process performs reliably under the intended conditions of use, e.g., external validation, subgroup analyses where relevant, and stress tests for foreseeable distribution shifts; (iv) \emph{uncertainty and calibration discipline}, i.e., whether uncertainty is represented and communicated in a manner that supports appropriate uptake, e.g., calibrated confidence estimates, abstention policies, or ``do-not-use'' triggers; (v) normative performance trade-offs as epistemic constraints, i.e., whether the chosen operating point, such as a sensitivity vs.\ specificity trade-off, reflects domain standards for acceptable error; and (vi) \emph{fairness of the target construct and its operationalization}, namely whether the task definition, label choice, proxy variables, and feature set are adequate given the domain's normative framework, including whether the chosen fairness criterion (e.g., predictive parity, equalized odds, calibration within groups) is epistemically appropriate for $\tau$ and whether fairness-accuracy trade-offs at the AI-alone level are coherently addressed at the PT-HAI level \citep{Donahue2022FAccT}.

\vspace{1em}

\noindent \texttt{type-RI$_3$} \emph{and socio-technical practices.} \texttt{type-RI$_3$} indicators capture the practices, institutional arrangements, and governance mechanisms that stabilize (or undermine) the reliability of the PT-HAI over time. Even where a human-AI team can in principle achieve strong \texttt{type-RI$_1$} performance and meet \texttt{type-RI$_2$} standards, reliability of the corresponding PT-HAI  may fail in practice unless there are durable procedures for competent use, monitoring, and change management. Accordingly, \texttt{type-RI$_3$} includes indicators such as: (i) \emph{training and competence scaffolding}, e.g., targeted education about failure modes, appropriate override behavior; (ii) \emph{workflow integration}, e.g., clear role assignment for $X$ and $A$, decision checkpoints, documentation of override; (iii) \emph{monitoring and feedback}, e.g., post-deployment performance tracking, incident reporting, auditing of systematic errors; (iv) \emph{accountability and incentives}, e.g., who is responsible for outcomes, whether institutional incentives encourage critical thinking; (v) \emph{lifecycle and update governance}, e.g., versioning,  communication of changes to users, and ensuring that the human’s expectations remain aligned with the system’s actual behavior.  \texttt{type-RI$_3$} indicators make explicit that the reliability of the PT-HAI is a property of an evolving socio-technical practice, providing evidence that the PT-HAI remains a reliable  computational process across time; and (vi) \emph{fairness audits and disparity-mitigation governance}, namely the institutional practices through which subgroup performance is monitored, disparities are surfaced and contested, remediation is triggered, and accountability for fairness outcomes is assigned.

\vspace{1em}

In summary, complementarity is an indicator of the reliability of PT-HAIs in the framework of computational reliabilism. It provides strong evidence for PT-HAI reliability when complementarity gains are non-trivial in magnitude (i.e., $\text{CTP}_\tau(D)=1$ is practically meaningful for the domain), stable across time and robust to foreseeable distribution shifts and system updates, and efficient, i.e., not achieved by shifting disproportionate cognitive or organizational burden onto human operators. However, complementarity is not enough to secure high levels of reliability by itself. In fact, as previously noted, complementarity is a purely relative performance criterion and a human-AI team can satisfy CTP even when the AI system is poorly reliable or the AI is highly reliable but the human contributor is not, e.g., an untrained user, so that the interaction remains fragile or unsafe despite occasional gains.\footnote{For instance, let $y_i^H = y_i - \varepsilon$, $y_i^{AI} = y_i + 0.5\varepsilon$, and $y_i^{HAI}=\tfrac{1}{2}(y_i^H+y_i^{AI})=y_i-0.25\varepsilon$ with $\varepsilon>0$. Then $L_{HAI}<\min\{L_H,L_{AI}\}$ even though all predictors are arbitrarily inaccurate.}
\textbf{High PT-HAI reliability therefore requires complementarity together with additional indicators of epistemic adequacy.} 
 Accordingly, this high reliability must also be supported by established validation practices, appropriate incentive structures, transparent communication and documentation, institutional accountability, and community-endorsed criteria of success. Since we introduced these indicators already, we close this discussion here.

\begin{figure*}[t]
  \centering
\includegraphics[width=\linewidth]{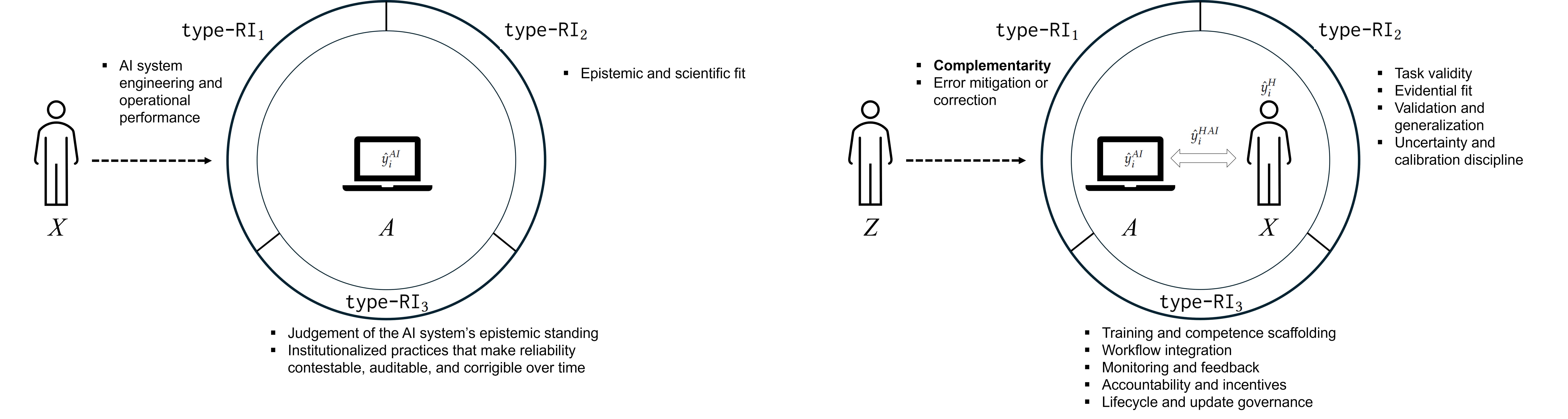}
  \caption{Two justificatory perspectives under CR. 
  \textbf{Left:} justification targets the AI system as a computational process in isolation as in Def.~\ref{def:CR_algo}. 
  \textbf{Right:} justification targets the PT-HAI as a computational process, where reliability emerges from interaction protocols, user competence, and organizational setting---see Def.~\ref{def:CR_hybrid}. 
  The same families of reliability indicators (type-RI$_1$-type-RI$_3$) apply to different targets depending on the justificatory perspective.}
\label{fig:two_justificatory_perspectives}
\end{figure*}

\subsection{A Brief Summary on the Two Justificatory Perspectives at Hand}
\label{subsection:two-perspectives}
We briefly clarify the two  justificatory perspectives discussed so far. We summarize them graphically in Figure \ref{fig:two_justificatory_perspectives}. In the first justificatory AI perspective---see  Section~\ref{subsection:CR_primer}---an agent \(X\), e.g., a data scientist using a model, or a clinician using a medical AI is justified in accepting the output of an AI system \(A\) as epistemically adequate for a task \(\tau\) iff \(A\) is reliable for \(\tau\) as in Def.~\ref{def:CR_algo}. This is the classical target of the justificatory AI literature \citep{duran2018grounds,Ferrario2024SEE}.
Our second perspective, see Def.~\ref{def:CR_hybrid}, shifts the unit of evaluation from the AI system to the PT-HAI. Here, a third party \(Z\), e.g., a patient, a manager, or a regulator, is justified in accepting a human-AI team output \(\hat y^{HAI}\) insofar as it is produced by a PT-HAI that is reliable for \(\tau\). Thus, what is epistemically justified is the output of a PT-HAI whose interaction protocol   determines how the human and the AI contribute to \(\hat y^{HAI}\). This second perspective is novel; it has been hinted at in the context of trust in human-AI  dyads by \citet{ferrario2022explainability}, but has not yet been developed in detail in justificatory AI.

A final question follows naturally: \emph{can a PT-HAI be reliable even if the AI system, considered on its own, is not reliable in the sense of Def.~\ref{def:CR_algo}?} On our account, the answer can be yes. An AI system may be an unreliable stand-alone predictor for \(\tau\) and still contribute to a reliable PT-HAI when the interaction protocol and context of use limit its influence on the final team output. For instance, low AI reliability can trigger scrutiny, enforce escalation and review that eventually result in complementarity instances,  increasing the degree of reliability of that PT-HAI. 

\subsection{How Does CR Address the Key Challenges of Complementarity?} 
Finally, we can return to the four challenges in Section~\ref{subsection:compl_challenges}. Our claim is not that CR solves complementarity. Rather, it repositions it, explaining what complementarity is good for, why its empirical fragility is not fatal, and what else must be measured and governed for PT-HAIs in everyday life to be epistemically defensible.

\vspace{1em}

\noindent\textbf{I. What theoretical anchoring for complementarity?} Rather than seeking a theory of when complementarity should emerge from information-processing models that are specific neither to AI systems, nor to different types of human-AI interactions, CR situates complementarity in an account of process reliability and epistemic justification. Within CR,  CTP matters because it can function as a reliability indicator: historical instances of complementarity provide defeasible evidence that the human-AI interaction protocol yields epistemic gain beyond either component alone, and thus that the interaction is a reliable process for task $\tau$ in context. 
Furthermore, instead of searching for a single, domain-general mechanism of complementarity, CR  encourages researchers to identify which interactional and socio-technical features strengthen the overall reliability profile \emph{in a given domain}, treating complementarity as one (potentially context-sensitive) signal within a family of reliability indicators.

\vspace{1em}

\noindent\textbf{II. Limited applicability of complementarity at decision time.}  
Complementarity remains an ex post metric in CR, but it is treated as historical evidence for assessing the reliability of a PT-HAI. Accordingly, it is helpful at decision-time as justification grounded in the dispositional reliability of the process, supported by past performance and other reliability indicators, rather than by a live estimate of $\text{CTP}_\tau(\cdot)$ for the current instance. In CR, the value of complementarity lies in its stability over time and in what it reveals about the drivers of effective teaming, not in serving as a decision-time property of any single interaction instance. CR also treats update-sensitivity as part of the reliability profile of the PT-HAI: if model updates, distribution shifts, or human learning undermine stability, this weakens relevant \texttt{type-RI$_1$} and \texttt{type-RI$_3$} indicators, such as monitoring, versioning, and communication practices. 

\vspace{1em}

\noindent\textbf{III. More than predictive accuracy is at stake in human-AI interactions.} By embedding complementarity inside a framework that already includes non-relative-performance desiderata via different types of reliability indicators, CR prevents CTP from being treated as a (relative) accuracy-only gold standard. CR clearly states that complementarity is  epistemically attractive \emph{only conditionally}, namely when it is achieved under constraints expressed by other reliability indicators of system robustness, human expertise, and institutional fit. Conversely, CR makes room for partially reliable PT-HAIs with limited complementarity when other reliability indicators are strong, e.g., where the key epistemic work is done by accurate humans under stringent validation, careful scoping of use, and robust governance that prevents overclaiming and out-of-scope deployment.

\vspace{1em}

\noindent\textbf{IV. Complementarity ignores the magnitude-cost profile of epistemic gain.}
CR does not treat all complementarity improvements as equally significant. Because reliability comes in degrees and depends on the interaction of multiple indicators, small performance gains achieved at high interactional or institutional cost may carry little justificatory weight, while larger or more robust gains may significantly strengthen the reliability profile of the process. On this view, complementarity is informative insofar as it contributes---relative to its costs and in combination with other indicators---to the overall reliability of the human-AI interaction. This cost-sensitive perspective in measuring efficient complementarity can be operationalized, as we will discuss in Section \ref{section:recommendations} and show in Section \ref{sub_app:complementarity_gain} in the Appendix. 

\subsection{Situating our Approach Within AI  Epistemology}
\label{subsection:approach_phil_AI}
Our contribution is not a general theory of epistemic justification for AI-assisted decision-making. Instead, we adopt CR to reposition complementarity within justificatory AI, clarifying what it can do in providing reasons to believe in the adequacy of AI-assisted predictions in real-world settings where humans interact with this technology. We believe CR offers a bridge between evidence on performance, epistemic standards, and institutional practices without collapsing epistemic justification into accuracy or treating complementarity as the sole locus of normative significance in human-AI interactions. 
Furthermore, it allows clarifying the relation between concepts, such as reliance, complementarity, and reliability, that permeate the human-AI interaction and justificatory AI literature, but still lack a unified, cross-domain perspective \citep{grote2024reliability}.
That said, any such choice leaves open questions that we cannot fully address here. While several classic criticisms of CR have received rebuttal \citep{beebe2004generality,comesana2006well,bonjour1980externalist,Goldman1986Book}, more recent challenges are worth mentioning \citep{alvarado2026challenges}.
 For instance, worries about \emph{warrant transmission and credit assignment} arise. In our account, we require that reliability claims be supported by ample families of contestable indicators that can be reported in checklists to warrant the transmission of reliability---see Section \ref{section:recommendations} and \ref{sub_app:minimal} in the Appendix.  Furthermore, while \citet{alvarado2026challenges} emphasizes the epistemic importance of \emph{endogenous features} (internal properties of the computational system), our account does not deny their relevance \emph{tout court}: it treats access to internal properties as one (often optional) contributor within a broader reliability profile alongside protocol design, user competence, and governance. Finally, in response to \emph{error-related opacity} and the difficulty of diagnosing failures from the outside \citep{alvarado2026challenges}, our account makes monitoring, escalation, and update governance \texttt{type-RI$_3$} indicators, so that failures are anticipated, detectable, and actionable over time.

\section{Three Examples of Complementarity in Computational Reliabilism}
\label{section:examples}
We continue by elaborating on complementarity as reliability indicator in CR through three real-world examples.

\subsection{The technologist dermatologist} 
Consider an experienced dermatologist who routinely uses an AI system that classifies dermoscopic images as benign or malignant in a modern clinic. The clinician is an active technologist: they consult the system's outputs and uncertainty cues, review disagreement cases, ask colleagues or domain experts when predictions conflict with clinical knowledge, and routinely attend courses on AI-assisted dermatology. The AI system is from a well-respected company that has published many empirical studies in leading journals and documents design, implementation, and maintenance protocols of the AI extensively. Furthermore, the clinic organizes routine training courses where the accuracy of AI is compared to that of its dermatologists and human-AI teams. Over time, errors and cases of CTP are discussed, analyzing the system's failure modes to calibrate the clinician's prediction accordingly. The clinic, in turn, runs routine training and audits comparing the performance of the AI, its dermatologists, and their PT-HAIs. In CR terms, all three reliability indicator families  carry relevant  justificatory weight for any patient relying on the AI-assisted clinician's prediction: they are part of highly reliable PT-HAIs. Complementarity (\texttt{type-RI$_1$})  strengthens reliability only insofar as gains are non-trivial and achieved without excessive burden onto the clinician. Given the increasing expertise of the dermatologist, this burden decreases over time.

\subsection{All complementarity and no credentials makes Jack a reliable student}
Jack, a mathematics student, uses a free large language model (LLM) to solve exercises, especially computation-heavy ones. The system is only sometimes helpful: it produces plausible-looking mistakes, skips assumptions, and occasionally invents steps. There is no strong task-specific validation, no institutional endorsement, and no credible basis for treating the model, on its own, as a reliable source of correct proofs or calculations. What can still make the PT-HAI somewhat reliable  is complementarity. Jack routinely forces the model to share alternative computations and intermediate steps,  tests edge cases, and cross-checks against textbooks and lecture notes. When the LLM is wrong, Jack often detects and repairs; when Jack is wrong, the LLM's solution  sometimes triggers correction. In CR terms, justificatory weight is carried mainly by \texttt{type-RI$_1$} through a stable pattern of error detection, correction, and complementarity gains. By contrast, \texttt{type-RI$_2$} and \texttt{type-RI$_3$} remain weak, so the overall PT-HAI displays low reliability. A second student, Lloyd, has forgotten to do his homework and asks Jack for last-minute help. On our view, Lloyd has reasons to accept the output of the Jack--LLM interaction  because Jack's past performance exhibits stable complementarity under a learning protocol. These reasons could be stronger if the LLM had been extensively tested for mathematical tasks, endorsed through use in academic contexts by lecturers and students alike. But  the deadline is coming soon and Lloyd has little time left.

\subsection{When complementarity does not pay}
A forensic lab processes a high volume of cases using an automatic speaker recognition tool to compare short, noisy robbery calls with suspect interviews. The tool can output a likelihood ratio, but only within validated conditions, e.g., duration, language, no strong disguise. Under these operational conditions, the (computational process realized by the) tool shows high reliability. Clarice, a certified forensic examiner, applies a strict protocol by checking recording quality, detecting artifacts, and documenting her assessment for court. She achieves high predictive accuracy consistently. Here, complementarity gains are  often marginal: Clarice and the tool usually agree in most cases that fall within validated conditions, and the human-AI team usually improves only slightly over the better component. However, the cost of this PT-HAI is important. Each extra iteration consumes  Clarice's scarce time and must be documented to a standard that survives legal and technical cross-examination, and explained in a way that is defensible in court. Small gains usually translate into relevant operational penalties, including delayed investigations, postponed hearings, and  growing court backlogs. 
The forensic lab treats certified cases of Clarice-tool complementarity as  welcome in those cases where operational costs are limited, but not a target to optimize. Reliability of the forensic lab investigation is secured mainly by \texttt{type-RI$_3$} constraints—validated scope, abstention triggers, peer review, proficiency testing, version control, and reporting conventions—precisely because they keep the process stable under high throughput and expert scarcity; this indicator family gives reasons to judges and other legal professionals to endorse the lab's outputs. Thus, on a CR view, this PT-HAI can be highly reliable even when complementarity gains are small, because what matters is not squeezing out marginal accuracy, but maintaining a defensible and court-robust epistemic practice.

\section{Recommendations for Design, Governance and Future Research}
\label{section:recommendations}
Our conceptual repositioning of complementarity  changes what complementarity is evidence for, how failures to achieve it should be interpreted, and what else must be measured and governed for AI-assisted decision-making to be epistemically defensible. 
We summarize implications of our approach for design and governance, operationalizing them through an assurance checklist and measures of efficient complementarity.

\subsection{Recommendations for human-AI design} 
Designers should not optimize for complementarity alone. Instead, they should treat it as one diagnostic signal within a broader reliability profile and report, at minimum, whether complementarity occurs, how large the gain is, how stable it remains over time, and what interactional and organizational costs are required to sustain it. 
In practice, this means prioritizing interaction protocols that make epistemically relevant behavior observable and correctable---for instance, structured contestation, escalation pathways, and explicit disagreement handling---rather than relying primarily on explanations or confidence displays that do not reliably improve teaming \citep{Vaccaro2024NHB}.
Finally, evaluation should report whether complementarity occurs and the magnitude, stability over time, and cost of the gain, since these dimensions determine its justificatory weight in CR.
Practically, human-AI interaction design should use a minimal assurance checklist that makes PT-HAI reliability assessable  by operators, governance and policy organizations, and affected third parties, e.g., patients, managers, regulators, who otherwise lack direct epistemic access to the PT-HAI protocol and its evidence base. Table~\ref{tab:pt-hai-min-checklist} in Section \ref{sub_app:minimal} in the Appendix specifies such a \emph{minimal reporting checklist} that designers can treat as a deployment deliverable: it documents the interaction protocol, provides interaction-level performance evidence (including complementarity where applicable), and encodes lifecycle commitments (monitoring and update governance) that foster PT-HAI reliability over time. Finally, designers should target and report \emph{efficient complementarity} explicitly: if gains require sustained monitoring, longer deliberation, or extensive retraining, the design may merely shift epistemic burden onto users and erode the practical value of the gain. 
Section \ref{sub_app:complementarity_gain} in the Appendix introduces a measure of net complementarity gain that takes into account the cost of complementarity, together with three complementary methods (policy-anchored, analogy-based, and deliberative) for eliciting the institutional threshold $\lambda$ that governs whether complementarity counts as efficient in a given setting.

\subsection{Recommendations for AI governance and policy} 
Regulators and institutions should avoid mandating complementarity as a compliance target, because doing so encourages  optimization of a relative, binary metric while leaving the broader reliability of the socio-technical process underspecified. Governance should evaluate PT-HAIs as evolving socio-technical practices rather than as static performance snapshots. In high-stakes settings, this requires collecting evidence of reliability through lifecycle-sensitive obligations such as monitoring, incident reporting, update documentation, revalidation triggers, and competence maintenance, since distribution shift, model updates, and human learning can rapidly erode the historical basis for justification.
This lifecycle-sensitive view on human-AI team reliability aligns with regulatory logics already visible in high-risk AI governance, for instance in the EU AI Act's emphasis on post-deployment monitoring, technical documentation, and change management---see Article 3(20) and Annex IV, Section 9 \citep{EU_AI_Act_2024}. More generally, policy should enforce proportionality: epistemic gains attributable to human-AI interaction should be commensurate with the cognitive and institutional burdens needed to sustain them, and small or fragile gains should trigger tighter scoping and stronger oversight requirements.

\section{Conclusions}
\label{section:conclusions}
Complementarity remains a valuable concept in human-AI interaction research, but not in the role it is currently asked to play. When treated as a stand-alone, ground-truth-dependent, relative-accuracy criterion, it is theoretically under-anchored and empirically fragile. When repositioned within computational reliabilism, it acquires a clearer role: it functions as historically grounded evidence about whether a prediction-task human-AI interaction is a reliable epistemic process for a task. This theoretical reframing turns complementarity from a binary success label into one component of a broader, graded account of epistemic justification that must also attend to validity, uncertainty discipline, competence scaffolding, and lifecycle governance if AI-assisted decision-making is to be epistemically defensible in high-stakes settings. 

\section*{Acknowledgments}
The work of A.~Ferrario and A.~Facchini was partly conducted within the framework of the EUonAIR Centre of Excellence in Responsible AI and Education. Their work was partially supported by a grant from Movetia, funded by the Swiss Confederation

\section*{Appendix}
\label{section:appendix}

\subsection{Justificatory prediction-task human-AI interactions: A minimal checklist}
\label{sub_app:minimal}

Table~\ref{tab:pt-hai-min-checklist} displays a minimal reporting checklist for justificatory PT-HAIs, classified by reliability-indicator families.

\begin{table*}[h!]
\centering
\small
\setlength{\tabcolsep}{5pt}
\renewcommand{\arraystretch}{1.2}
\begin{tabular}
{>{\raggedright\arraybackslash\leftskip=1em}p{0.20\linewidth}
p{0.07\linewidth} p{0.62\linewidth}}
\hline \textbf{Item} & \textbf{\texttt{type-RI$_\bullet$}} & \textbf{What to document (minimum)} \\
\hline
\textbf{AI scope and conditions of use} & \texttt{RI$_2$}, \texttt{RI$_3$} & Intended use, in-/out-of-scope cases, assumptions about environment, and boundary conditions that trigger abstention/escalation. \\
\textbf{Protocol} $\Pi$ & \texttt{RI$_1$}, \texttt{RI$_3$} & How the human consults and integrates AI outputs; disagreement handling; escalation/second review; how $\hat y^{HAI}$ is produced. \\
\textbf{User competence} & \texttt{RI$_3$} & Who is authorized; training content and cadence; competence checks; known failure modes covered. \\
\textbf{Performance} & \texttt{RI$_1$} & $L_H$, $L_{AI}$, $L_{HAI}$ (or task-appropriate metrics). \\
\textbf{Complementarity evidence} & \texttt{RI$_1$} & CTP, gross gain $\Delta_\tau(D)$ and net gain $\Delta^{\text{net}}_\tau(D)$. Report magnitude and stability across time.  \\
\textbf{Interaction cost} & \texttt{RI$_3$} & Cost term $c(D)$ with clear definition and unit, e.g., minutes, staff-hours, \$. Include cost categories, e.g., review/second opinions, auditing, documentation, throughput loss and how costs are estimated. \\
\textbf{Efficient complementarity} & \texttt{RI$_1$}, \texttt{RI$_3$} & Net gain $\Delta^{\text{net}}_\tau(D)$ and whether it is positive. Report (1) the efficiency ratio $\Delta_\tau(D)/c(D)$, 
(2) the institutional threshold $\lambda$ (3)  the elicitation method used to set $\lambda$ and (4) the review cadence under which $\lambda$ is revisited.
\\
\textbf{Uncertainty discipline} & \texttt{RI$_1$}, \texttt{RI$_2$} &  When uncertainty in predicting triggers human review and how this uncertainty is communicated and acted upon. \\
\textbf{Epistemic validity} & \texttt{RI$_2$} & Why the target, labels, and features are appropriate for the decision purpose in the given epistemic context. \\
\textbf{Update and drift management} & \texttt{RI$_3$} & Versioning, remodeling and evaluation triggers, and change communication. \\
\textbf{Monitoring and accountability} & \texttt{RI$_3$} & Post-deployment tracking, incident reporting, auditing,  responsibility assignment. \\
\textbf{Subgroup performance} & \texttt{RI$_1$} & Disaggregated joint accuracy, error rates, and calibration across operationally relevant subgroups; comparison with AI-alone and human-alone disaggregated performance. \\
\textbf{Fairness of target construct} & \texttt{RI$_2$} & Justification for the chosen fairness criterion; analysis of label, proxy, and feature choices under that criterion; explicit treatment of fairness-accuracy trade-offs. \\
\textbf{Fairness audits and governance} & \texttt{RI$_3$} & Audit cadence, remediation pathways, accountability assignment, incident reporting for disparity events. \\
\hline
\end{tabular}
\caption{Minimal reporting checklist for justificatory PT-HAIs, classified by reliability-indicator families. The list is intentionally minimal; richer domains may require additional constraints. 
However, a justificatory claim without these basics is underspecified.}
\label{tab:pt-hai-min-checklist}
\end{table*}

\subsection{Measures of complementarity gain and its efficiency}
\label{sub_app:complementarity_gain}
We introduce a formalization of complementarity gain that captures (1) the magnitude of the predictive improvement achieved by a human-AI team, and (2) the interactional costs required to obtain and sustain that improvement. 

\subsubsection{Gross complementarity gain.}
Let $\tau$ be a prediction task and $D$ a dataset. Suppose a PT-HAI achieves complementarity on $D$, i.e., $\mathrm{CTP}_\tau(D)=1$. We define the \emph{gross complementarity gain} as:
\begin{equation}
\Delta_\tau(D) \;:=\; \min\{L_H(D),\,L_{AI}(D)\} - L_{HAI}(D),
\end{equation}
where $L_H$, $L_{AI}$, and $L_{HAI}$ are the empirical losses of the human, the AI system, and the human-AI team, respectively, computed under the task-appropriate loss function (Section~\ref{subsection:complementarity}). By construction,
$\Delta_\tau(D)>0$ if and only if $\mathrm{CTP}_\tau(D)=1$---see Def.~\ref{def:complementarity}. 
$\Delta_\tau(D)$ measures the absolute performance improvement of the PT-HAI relative to the better standalone component and has the same units as the loss. For instance, if $L(\cdot)$ is mean squared error, then $\Delta_\tau(D)$ has units $u^2$ (where $u$ is the unit of the target variable); if $L(\cdot)$ is cross-entropy, then $\Delta_\tau(D)$ is dimensionless.

\begin{figure}[h!]
\centering
\begin{tikzpicture}[scale=1.0, >=Latex, every node/.style={font=\small}]

\draw[->] (0,0) -- (6.2,0) node[right] {$c(D)$};
\draw[->] (0,0) -- (0,4.8) node[above] {$\Delta_\tau(D)$};

\draw[thick] (0,0) -- (5.2,3.1) node[pos=0.90, right] {$\Delta_\tau(D)=\lambda c(D)$};

\draw[dashed, thick] (0,0) -- (3.6,4.2);

\node[align=center] at (3.5,4.8) {\textbf{efficient}\\complementarity};
\node[align=center] at (3.5,1.0) {\textbf{inefficient}\\complementarity};

\fill (4.3,3.5) circle (1.4pt);
\node[right] at (4.3,3.5) {$\big(c(D),\Delta_\tau(D)\big)$};

\draw[->] (5.8,4.55) to[bend left=20] (3.2,3.7);
\node[align=left] at (6.8,4.8) {higher $\lambda$: \\ steeper threshold};

\draw[densely dotted] (4.3,0) -- (4.3,3.5);
\draw[densely dotted] (0,3.5) -- (4.3,3.5);

\end{tikzpicture}
\caption{Efficient complementarity in the $(c,\Delta)$ plane. Complementarity is efficient when the gross gain exceeds the threshold line, i.e., when $\Delta_\tau(D)>\lambda c(D)$. Points above the line represent efficient complementarity; points on or below it represent inefficient complementarity. A higher $\lambda$ yields a steeper threshold and therefore makes efficient complementarity harder to achieve.}
\label{fig:efficient_complementarity}
\end{figure}

\subsubsection{Net complementarity gain and efficiency.}
Achieving complementarity may require non-trivial interactional and organizational effort as discussed in Section \ref{subsection:compl_challenges}. We capture this through a cost term $c(D)$, which aggregates resources expended to sustain the PT-HAI over $D$. Depending on the domain, $c(D)$ may include monitoring and review effort, e.g., second opinions and audits, and organizational costs, e.g., documentation, escalation procedures, reduced throughput. Costs may be quantified in context-appropriate units, such as  minutes, staff-hours, or monetary cost.

Because predictive loss and interaction cost generally have different units, we introduce a conversion parameter $\lambda>0$ with units (loss units)/(cost units), and define the \emph{net complementarity gain} as:

\begin{equation}
\Delta^{\mathrm{net}}_\tau(D) \;:=\; \Delta_\tau(D) - \lambda\,c(D) \quad (\text{where }\Delta_\tau(D)>0).
\end{equation}

From the definition of net complementarity gain, it follows that it is positive $(\Delta^{\mathrm{net}}_\tau(D)> 0)$ when

\begin{equation}
\frac{\Delta_\tau(D)}{c(D)}>\lambda \quad (\text{where }c(D)>0).
\end{equation}

Then, we say that \textbf{complementarity on $D$ is efficient when it produces positive net gain $\Delta^{\mathrm{net}}_\tau(D)$, i.e., when the observed gain $\Delta_\tau(D)$ per unit of interaction cost exceeds a threshold $\lambda$}. Intuitively, $\lambda$ specifies the minimum rate of predictive improvement per unit of interaction cost that an institution requires for complementarity to count as worthwhile in a given setting.

On the one hand, higher $\lambda$ values correspond to settings, for instance, high-risk institutions such as hospitals, where interaction is treated as expensive as compared to, for instance, reliance. On the other hand, lower values correspond to settings where interaction is comparatively acceptable, e.g., training or low-stakes contexts. In summary, different institutions may legitimately adopt different values of $\lambda$ for the same task, depending on stakes, throughput constraints, personnel scarcity, and governance norms. For this reason, claims about efficient complementarity are not fully interpretable unless the chosen $\lambda$ is reported and justified.

Within CR, $\lambda$ functions as a \texttt{type-RI$_3$} parameter as it makes explicit a judgment about acceptable trade-offs between epistemic improvement and the resources required to sustain it. 
In practice, a value for $\lambda$ can be introduced using internal policies in a given social context. For instance, by implementing the rule ``we add one minute of expert review of AI predictions to achieve complementarity only if it reduces  loss by at least $\delta$ (loss units),'' one has $\lambda=\delta  ~\text{(loss units)}/\text{minute}$, where $\delta>0$. The policy-anchored example above is the simplest elicitation strategy but not the only one available. Two additional methods are relevant when no prior institutional rule exists or when richer justification is required. (I) \emph{Analogy to decision-theoretic precedents}: For instance, in medical settings, $\lambda$ can be calibrated against established willingness-to-pay thresholds from health-economic evaluation---e.g.,  QALY-based decision rules \citep{loomes1989use} re-expressed in the relevant loss unit such as false-negative reduction per staff-hour---whereas in forensic contexts, $\lambda$ can be anchored to cost-sensitive likelihood-ratio frameworks already used to weigh investigative delay against evidential gain. (II) \emph{Deliberative elicitation}:   For high-stakes deployments where stakeholder buy-in is itself part of the reliability profile, $\lambda$ can be set through structured consultation with domain experts, affected parties, and oversight bodies, producing a documented rationale that can subsequently be audited. Across all three proposed methods (policy-anchored, analogy-based, and deliberative), $\lambda$ is a governance artefact. Its value, its derivation, and the cadence of its review should appear in the minimal reporting checklist (see Table \ref{tab:pt-hai-min-checklist}).  Different methods will be appropriate in different domains, but what unifies them is that they render the institutional judgment about acceptable trade-offs explicit and contestable, in line with the type-RI, role $\lambda$ already plays in our framework.

Geometrically, the condition for efficient complementarity is $\Delta_\tau(D)>\lambda c(D)$. This means that, in the $(c,\Delta)$ plane, efficient cases lie above the threshold line with slope $\lambda$, while inefficient cases lie on or below it. Higher values of $\lambda$ make this threshold steeper and therefore make efficient complementarity harder to achieve. We show this in Figure~\ref{fig:efficient_complementarity}.

\bibliography{aaai2026}

\end{document}